\documentclass[letterpaper, 10 pt, conference]{ieeeconf}  

\IEEEoverridecommandlockouts                              

\usepackage[linesnumbered,ruled,vlined]{algorithm2e}
\usepackage{amsfonts, amsmath}
\usepackage{bm}
\usepackage{graphicx}
\usepackage{hyperref}

\SetKwInput{KWInput}{Input}
\SetKwInput{KWOutput}{Output}
\SetKwInput{KWData}{Data}

\graphicspath{ {./images/} }

\title{\LARGE \bf
Integrating Human Expertise in Continuous Spaces: A Novel Interactive Bayesian Optimization Framework with Preference Expected Improvement
}

\author{Nikolaus Feith$^{1}$ and Elmar Rueckert$^{1}$
\thanks{$^{1}$Chair of Cyber-Physical Systems,
        Montanuniversität, 8700 Leoben, Austria, \newline
        corresponding author: {\tt\small nikolaus.feith@unileoben.ac.at}}%
}

\begin{document}

\maketitle
\thispagestyle{empty}
\pagestyle{empty}

\begin{abstract}

Interactive Machine Learning (IML) seeks to integrate human expertise into machine learning processes. However, most existing algorithms cannot be applied to Realworld Scenarios because their state spaces and/or action spaces are limited to discrete values. Furthermore, the interaction of all existing methods is restricted to deciding between multiple proposals. We therefore propose a novel framework based on Bayesian Optimization (BO). Interactive Bayesian Optimization (IBO) enables collaboration between machine learning algorithms and humans. This framework captures user preferences and provides an interface for users to shape the strategy by hand. Additionally, we've incorporated a new acquisition function, Preference Expected Improvement (PEI), to refine the system's efficiency using a probabilistic model of the user preferences. Our approach is geared towards ensuring that machines can benefit from human expertise, aiming for a more aligned and effective learning process. In the course of this work, we applied our method to simulations and in a real world task using a Franka Panda robot to show human-robot collaboration.

\end{abstract}

\section{INTRODUCTION}
Interactive Machine Learning (IML) is a field within artificial intelligence that seeks to incorporate human expertise directly into machine learning processes via a feedback loop. This approach is notably suitable for reinforcement learning (RL) scenarios. Two primary concerns arise within IML: First, there is a challenge of presenting the machine learning model to a human decision maker (DM). Second, what's the best method to convey human feedback to the algorithm, and what type of input is most beneficial? It's essential to address these questions, especially considering the different ways humans and machines interpret data. Efforts should be made to bridge this understanding gap, ensuring a more harmonious interaction between the two.
\begin{figure}[thpb]
  \centering
  \includegraphics[scale=0.50]{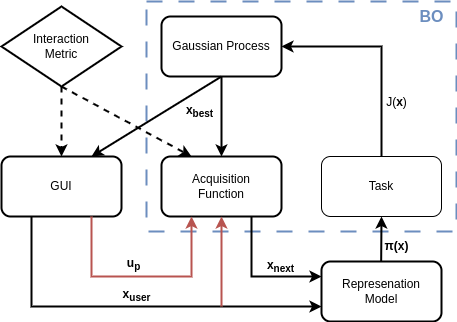}
  \caption{Shows the architecture of Interactive Bayesian Optimization. The black arrows indicate the baseline data flow, the red arrows are additionally necessary for PEI. In blue BO is marked without interaction and representation model.}
  \label{fig:IMLfigure}
\end{figure}
In our research, we're taking a new approach to address the complex questions we've identified. We view our reinforcement learning challenge as an optimization task, where humans and algorithms collaborate to refine the agent's actions. Our research introduces a novel framework, grounded in Bayesian Optimization, which facilitates a smoother integration of human feedback into reinforcement learning. Central to our approach is a unique acquisition function, designed to enhance the efficiency of the learning process and better capture user preferences. A distinctive feature of our framework is its dual capability: it gathers user preferences and also permits users to adjust the RL agent's policy via an intuitive graphical interface. We've further enhanced our system with a new acquisition function named 'Preference Expected Improvement', building upon the foundational 'Expected Improvement'(EI) concept. Anticipating real-world robotics applications, our system is developed as a ROS2 application. Data flow between Bayesian Optimization, RL, and our web interface is managed through ROS2 Topics. This modular design offers flexibility to adapt the components to the specific problem.

Very close related to our method is Preference Reinforcement Learning \cite{a1, a2, a3}, where in each iteration two or more solutions are shown to the user and the user chooses one. Based on this, the preferences are learned and it is adjusted which new solutions are generated. For this, the preference must be considered as a probabilistic model as shown in \cite{a4, a5}. The goal is to guide the exploration in such a way that it proceeds in the sense of the user. In contrast, we show the DM the current best policy, which is then modified by the decision maker and the user's preferences are queried. Consequently, the human gradually changes the best solution, so that the policy is shaped progressively to the user's ideas.

Initially, we'll explore existing interactive machine learning frameworks, highlighting their relevance and connection to our approach. Subsequently, we'll introduce our framework and delve into the experiments we've conducted, analyzing the outcomes. Concluding our study, we'll provide definitive answers to our research questions.

\section{RELATED WORK}
In our research, we narrow our focus to the application of IML \cite{a6,a7} specifically to RL challenges called Interactive Reinforcement Learning (IRL). While there are numerous ways to categorize Interactive Machine Learning algorithms, we will classify them based on real-time interactions and episodic interactions.
\subsection{Real time interaction}
Real-time interaction methods involve immediate user responses, typically within a span of 0.1 to 4.0 seconds, following the agent's action. The interaction is typically binary, with inputs of (+1, -1). Notable implementations of this method are found in Policy Shaping (PS) \cite{a8}, TAMER \cite{a9}, and CAIR \cite{a10}. These methods are based on various reinforcement learning techniques. For instance, PS employs Bayesian Q-learning to discern the best policy. TAMER, on the other hand, sidesteps the conventional Markov Decision Process reward function, instead striving to emulate the human user's internal reward mechanism. Given that both TAMER and PS operate exclusively in discrete state and action domains, extensions like ContinuousTAMER \cite{a11}, Tamer+RL \cite{a12}, DeepTAMER \cite{a13}, and DQNTAMER \cite{a14} have been introduced to accommodate continuous state-spaces. CAIR, in contrast, leverages the Soft Actor Critic architecture to learn both internal and environmental reward functions, proving effective even in high-dimensional continuous realms.

We are convinced that a direct comparison of IBO and the real time methods is not useful, because our method queries the user episodically. Furthermore, with the exception of CAIR all methods are based on a discrete action space. CAIR is based on a very high frequency of user interactions. In their Human Subjects Validation, the performance was evaluated in a slightly slowed down simulation where ~4-5 interactions took place every second in a maximum period of 60 seconds. Since only binary feedback can be given and the sheer number of interactions, a comparison with Interactive Bayesian Optimization has few value. 

\subsection{Episodic interactions}
For episodic interactions, entire episodes are shown to the human and then the interaction takes place. COACH \cite{a15} is based on the Actor-Critic architecture, using episodic interactions with the possibility to change to real interactions. The algorithm shows a policy to the user and in addition the policy with on action changed. Now the users decides if it a positive, neutral or negative change. COACH was extended with DeepCOACH \cite{a16}, so that the raw pixels are used and not the state of the agent. Meanwhile, other works examine the challenge of utilizing pairwise preferences in IRL \cite{a2, a17, a18, a19}. These preferences, often elicited from demonstrations or queries, offer a more nuanced understanding of desirable behavior without relying on explicit reward values. Recent advancements also highlight the application of preference-based IRL in real-world scenarios, from robotic tasks to complex decision-making environments \cite{a20, a21}. As this field continues to evolve, the integration of human preferences promises to make reinforcement learning agents more aligned with human values, adaptable, and capable of generalizing across diverse tasks.

In our work, we concentrate on episodic interaction in a robotics application. Therefore, action and state space must be continuous. In our focus is the possibility of full interaction, so the human user can change the complete policy and is not limited to decide which of the suggestions is preferable. In this sense it is a different to the preference reinforcement learning methods and overcomes the pure preference learning with the possibility to change the policy directly.

\section{METHODS}
\label{sec: methods}
Our framework is based on Bayesian Optimization which is briefly introduced in this section. Subsequently, our Preference Expected Improvement Acquisition function is presented and finally, the structure of our method is discussed.

\subsection{Problem Statement}
In our research, we conceptualize the reinforcement learning environment using a Markov Decision Process (MDP). This MDP is characterized by the tuple $(S,A,T,R)$, wherein $S$ enumerates the entirety of potential states, $A$ comprises the full spectrum of actions, $T$ articulates the transition function as $T: S \times A \to Pr[S]$, and $R$ stands for the reward function denoted by $R: S \times A \to \mathbb{R}$. The overarching aim is to identify the optimal policy $\pi^*(\bm{\theta}, s(t))$ which, for any given state $s(t)$,elects an action $a(t)$ in pursuit of maximizing the cumulative reward or return
\begin{equation*}
   J(\bm{\theta}) = \mathbb{E}\left[ \sum_{t=0}^{T}{\left\{ r_t(s(t), a(t)) | \pi(\bm{\theta},s(t)) \right\} } \right]. 
\end{equation*}
Throughout this investigation, we have adopted a deterministic policy $ \pi(\bm{\theta},s(t)) $, designed to yield a distinct action $a(t)$ for every state $s(t)$. This is achieved using a Representation Model (RM), computing the policy from $\bm{\theta}$. The learning process for this policy vector necessitates the establishment of correlations between $ \bm{\theta} $ and the expected return. This is performed through Gaussian process modeling, followed by the generation and subsequent evaluation of samples. The observations utilized are obtained from a Bayesian Optimization (BO) algorithm. The central optimization challenge:
\begin{equation*}
    \boldsymbol{\theta}^* = \operatorname{argmax}_{\boldsymbol{\theta}} \mathbb{E} \left[ J(\boldsymbol{\theta})| \pi(\boldsymbol{\theta})  \right],
\end{equation*}
seeks an efficient sample-based optimal solution.

\subsection{Bayesian Optimization}
Bayesian Optimization \cite{a22, a23} stands as a probabilistic approach tailored for the global optimization of expensive black box functions. Its inherent strength lies in its capacity to pinpoint a reasonably precise solution with minimal iterations. Given the costly nature of evaluating these black box functions, an acquisition function plays a pivotal role in guiding the selection of the subsequent optimal solution for evaluation. One limitation is the number of dimensions of the input space \cite{a24}. Starting with ten dimensions its performance decreases significantly and above 20 dimensions the search space is too vast to optimize in a few hundred steps to use it in interactive learning. As we will show in section \ref{sec: Results}, this limitation can be overcome with the use of IBO. Subsequent sections will delve deeper into the foundational concepts standard BO.

\subsubsection{Gaussian Process}
Bayesian Optimization predominantly employs Gaussian Processes (GP) as the underlying model to map an input vector $x \in X$ to an output scalar $y \in Y$. In the context of our research, this model translates the parameter vector $\bm{\theta}$ to the cumulative return $J(\bm{\theta})$. The overarching objective is to harness the acquisition function in guiding the selection of the ensuing observation, which in turn refines the model. The Gaussian process, as described in Equation (1), distinguishes between the data $D = \{X,y\}$ observed thus far, and $ D_* = \{X_*, y_*\}$, the prospective query.
\begin{equation*}
    \begin{bmatrix}
        y \\
        y_*
    \end{bmatrix}
    \sim 
    \mathcal{N} \left(
    \begin{bmatrix}
        m(X))  \\
        m(X_*)
    \end{bmatrix},
    \begin{bmatrix}
        K(X,X) & K(X,X_*) \\
        K(X_*,X) & K(X_*,X_*) 
    \end{bmatrix} \right). \tag*{(1)}
\end{equation*}
The GP is wholly characterized by its mean function $\mathbf{m}(X)$ and its covariance function, also referred to as the kernel $\mathbf{K}$. We initialize the mean function to $\mathbf{m}(X) = 0$ and select the Matern kernel,which performed best among numerous alternatives, as represented in Equation (2), for the covariance function.
\begin{equation*}
    k_\nu(\mathbf{x}_p, \mathbf{x}_q) = \sigma^2 \frac{1}{2^{\nu - 1} \Gamma(\nu)} \mathbf{A}^\nu \mathbf{H}_\nu \mathbf{A} + \sigma^2_y \delta_{pq}\ . \tag*{(2)}
\end{equation*}
This kernel serves as an extension of the quadratic exponential kernel, given the Matern kernel's inclusion of a smoothing parameter $\nu$. For our purposes, we adopt the Matern kernel with $\nu = 1.5$. Here, $\Gamma$ represents the gamma function, $\mathbf{H}_\nu$ denotes the modified Bessel function, $l$ specifies the kernel's width, $A$ is defined as $A = \left(\frac{2 \sqrt\nu \left\| \mathbf{x}_p - \mathbf{x}_q \right\|}{l} \right)$, and $\delta_{pq}$ is the Kronecker delta function. The Gaussian Processes in our methods are refined leveraging the marginal likelihood.

\subsubsection{Acquisition Functions} \label{subsubsection: Acq}
Central to Bayesian Optimization (BO) is the determination of subsequent input values that should be queried to either enhance the model or diminish its variance, thereby aiding in the discovery of an optimal solution. This is where acquisition functions come into play.
\begin{align*}
    \mathbb{E}(y_*|\mathbf{y}, X, \mathbf{x}_*) &= \mu(\mathbf{x}_*) = m_* + \mathbf{K}_*^\top \mathbf{K}^{-1}(\mathbf{y}- \mathbf{m})\ , \tag*{(3)} \\
    \text{Var}(y_*|\mathbf{y}, X, \mathbf{x}_*) &= \sigma(\mathbf{x}_*) = \mathbf{K}_{**} + \mathbf{K}_*^\top \mathbf{K}^{-1}\mathbf{K}_* \ . \tag*{(4)}
\end{align*}
 With Equations (3) and (4), we can sample new input vectors utilizing the surrogate model $x \sim \mathcal{N}(\mu(x_*), \sigma(x_*)$. Given the vast computational cost of evaluating the black-box function for every conceivable input vector, the acquisition function steps in to identify the most promising sampled vector. Among the countless acquisition functions available, we will spotlight the three most popular ones, that we have also employed: Expected Improvement $a_{EI}$, Probability of Improvement $a_{PI}$, and Upper Confidence Bound $a_{UCB}$, detailed in Equations (5) through (7).
\begin{align*}
    a_{\text{EI}}(x;\mathcal{D}) &= \mathbb{E}[\max  \mu_{\mathcal{D}'}(\mathbf{x})| x, \mathcal{D}] - \max \mu_{\mathcal{D}}(x) \\
    &= (v) \mathbf{\Phi}\left( \frac{v}{\sigma} \right) +  \sigma \phi\left( \frac{v}{\sigma}  \right) \ ,\tag*{(5)} \\
    a_{\text{PI}}(x;\mathcal{D}) &= \mathbf{\Phi}\left( \frac{v}{\sigma} \right) \ ,\tag*{(6)}\\
    a_{\text{UCB}}(x;\mathcal{D}) &= \mu(x) + \lambda \sigma(x) \ . \tag*{(7)}
\end{align*}
Each function uniquely balances the trade-off between exploration and exploitation. Here, $\Phi$ denotes the cumulative distribution function, $\phi$ is the density function and with $v = \mu - f(x^*) - \kappa$, and $\lambda, \kappa$ are hyper parameters. Utilizing the acquisition function, the optimal candidates from the sampled input vectors can be identified for evaluation in the black-box function. Subsequently, the surrogate model is updated using Equation (1), enabling the sampling of the next set of input vectors. It's important to note that the new vectors are typically sampled from a uniform distribution. However, none of the acquisition function considers human feedback. To integrate human feedback we propose a novel acquisition function called Preference Expected Improvement.

\subsection{Preference Expected Improvement (PEI)} \label{subsubsection: PEI}
The PEI represents our contribution to acquisition functions, tailored specifically for IML. At the heart of this method lies the modification of the distribution from which new input vectors are sampled from. As mentioned before, a uniform distribution is utilized, or given adequate knowledge regarding the input search space, more tailored distributions are employed. In our approach, we use a normal distribution as the prior, characterized by a notably large variance. To ensure that only those samples within the boundaries of the input search space are considered, we employ rejection sampling. Given the extensive variance and the limited search space, the resultant samples approximate a uniform distribution, called the proposal distribution. Subsequently, the EI acquisition function guides the selection of the next observation and evaluated in the simulation. This is the non-interactive optimization using Bayesian Optimization.
\begin{align*}
    \bm{\mu}_{pref} &= \bm{x}_{best} + \Delta\bm{x}_{user}\ , \tag*{(8)} \\
    \Sigma_{pref} &= diag(\sigma_1,\sigma_2,...) \ , \tag*{(9)}\\
    \text{with }\sigma_i &=
    \left\{
    \begin{array}{ll}
    \sigma_i = \sigma_0 & \text{if} \ u_i = false \\
    \sigma_i = \sigma_{pref} & \text{if} \ u_i = true 
    \end{array}
    \right. \tag*{,}\\
    p(\bm{x}_{user},\bm{x}) &= p(\bm{x}_{user}|\bm{x}) \ p(\bm{x}) \ , \tag*{(10)}
\end{align*}
To refine this proposal distribution, we ask the user in the GUI for the next input vector $\bm{x}_{user} = \bm{\mu}_{pref}$, Equation (8), as well as whether the respective dimension should be further explored or the selected value is preferred. The preference information is stored in the $u_{pref}$ array and is used to adjust the variance, Equation (9), of the likelihood $p(\bm{x}_{user}|\bm{x}) = \mathcal{N}(\bm{x}_{user}|\bm{\mu}_{pref}, \Sigma_{pref})$, where $\bm{\mu}$ is unknown. Subsequently, the conditional joint distribution, Equation (10), is calculated with the prior, i.e. the proposal distribution $p(\bm{x}) = \mathcal{N}(\bm{x}|\bm{\mu}_{prop}, \Sigma_{prop})$. Since all distributions are Gaussian distributions we use the closed-form to compute the posterior $p(\bm{x}|\bm{x}_{user}) = \mathcal{N}(\bm{\mu}_{post}, \Sigma_{post})$ i.e. the new proposal distribution:
\begin{align*}
    p(\bm{x}|\bm{x}_{user}) &\propto  p(\bm{x}_{user},\bm{x}) \ ,  \\
    \Sigma_{post} &= (\Sigma_{prop}^{-1} + \Sigma_{pref}^{-1})^{-1} \ , \tag*{(12)} \\
    \mu_{post} &= \Sigma_{post} \ (\Sigma_{prop}^{-1} \ \mu_{prop} + \Sigma_{pref}^{-1} \ \mu_{pref}) \ . \tag*{(13)}
\end{align*}
Within this algorithm, parameters that are designated as preferred exhibit a constrained search range, while the remaining parameters maintain an approximately uniform distribution. This structure empowers users to precisely dictate which parameters should be more explored and which are sufficient enough limiting the search range.

\subsection{Interactive Bayesian Optimization}
Interactive Bayesian Optimization has a modular architecture consisting of four components, which are shown in Fig. 1 
and described in Algorithm 1. The core of IBO is a Bayesian Optimization with a GP using Matern kernel as surrogate model and our novel PEI as acquisition function. The surrogate model takes input vector $\bm{\theta}$, which is denoted as $\bm{x}$ in the following, and corresponding $J(\bm{\theta})$ denotes the return $J(\bm{x})$ of the RL environment. 

At the beginning of each episode the Interaction Metric (IM) is called to decide whether the user should be queried. Throughout our experimental analysis, we deployed three distinct interaction metrics:

\begin{itemize}
  \item The random metric, inspired by the $ \epsilon $-greedy action selection strategy, entails drawing a random number within the $ [0;1] $ interval. If this value surpasses the $ \epsilon $ threshold, human input is queried.
  \item The regular metric consistently seeks user input at predefined intervals.
  \item The improvement metric computes the improvement rate over a fixed number of recent iterations and, if this rate falls below a threshold, the decision-maker's input is asked.
\end{itemize}

For a non-interactive iteration, the next input vector $\bm{x}$ is selected using the standard BO procedure. The representation model takes $\bm{x}$ and computes the policy, which is evaluated in the RL environment. Subsequently, the surrogate model is updated. In the case of an interactive episode, the current best policy is shown to the GUI. Next, the DM can modify the parameters Equation (8), set the DM's preferences Equation (9), or do a mixture of both. As with the non-interactive, the policy is calculated, evaluated, and the GP updated. Finally, as described in \ref{subsubsection: PEI}, the proposal model is improved with Equations (12),(13) and the next episode begins.
\begin{algorithm}
\caption{Interactive Bayesian Optimization}
\KWData{\\nr\_episodes, \\ proposal distribution $\mathcal{N}(\mathbf{\mu}_{prop}, \mathbf{\Sigma}_{prop})$}
\For{nr\_episodes}{
$m \leftarrow$ metric.\\
\If{$\neg m$}{
    $\mathbf{x} \sim$ rejection\_sampling( $\mathcal{N}(\mathbf{\mu}_{prop}, \mathbf{\Sigma}_{prop})$).\\
    $\mathbf{x}_{next} \leftarrow max(a_{PEI}(\mathbf{x}, \mathcal{D}))$.\\
    $y =$ model\_evaluation$(\mathbf{x}_{next})$. \\
    update\_GP$(\mathbf{x}_{user}, y)$ 
}\Else{
$\mathbf{x}_{best} \leftarrow \operatorname{argmax}_y(\mathcal{D})$.\\
display $\mathbf{x}_{best}$ in user interface.\\
$\mathbf{x}_{user}, u_{p}\leftarrow$ user interaction.\\
$y =$ model\_evaluation$(\mathbf{x}_{user})$.\\
update\_GP$(\mathbf{x}_{user}, y)$. \\ 
$\mathbf{\mu}_{prop}, \mathbf{\Sigma}_{prop} \leftarrow$ update\_proposal$(\mathbf{x}_{user}, u_{p})$.
}
}
\end{algorithm}
\section{RESULTS} \label{sec: Results}
To evaluate our framework, we conducted three experiments, i.e. the 'Cartpole Balancing' environment from the OpenAI gym library \cite{a24}, the 'Reacher' environment from DeepMind Control Suite \cite{a25} to test two dimensional environments and finally, we learned optimal reaching policies using a real Franka Emika Panda robot. As base lines we chose the results of the framework without interaction, using the three standard acquisition functions: UCB, PI, EI, where in this results only EI is listed all other base line results can be found in the supplementary pages: \url{https://sites.google.com/view/interactive-bo}.

In both simulated environments, we performed three under experiments each:
\begin{enumerate}
    \item[a)] Preference: User does not modify the individual parameters, but determines which of the parameters are preferred, Equation (9).
    \item[b)] Shaping: DM adjusts the individual parameters, but does not specify any preferences, Equation (8).
    \item[c)] Mixture: User can adjust the parameters as well as specify the preferences, Equation (8)+(9).
\end{enumerate}
The hyper parameters used can be found in the supplementary pages. We used a Gaussian basis as a representation model for the experiments to better evaluate the performance depending on the number of dimensions and type of action spaces.
\begin{figure*}
    \centering
    \includegraphics[width=\textwidth]{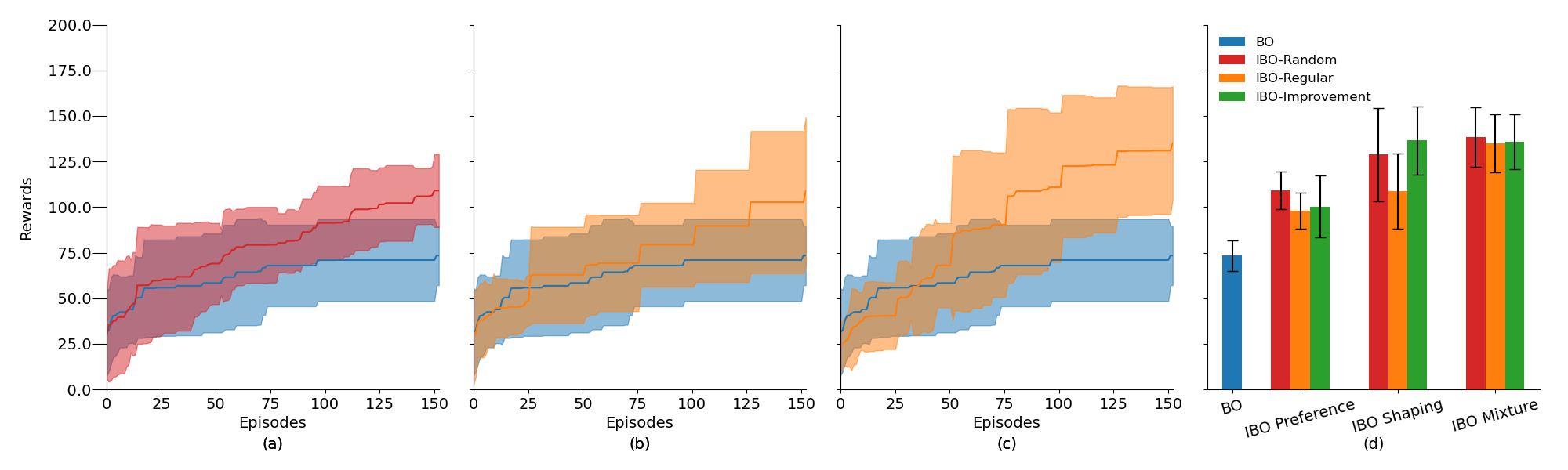}
    \caption{Shows Cartpole results for (a) preference, (b) shaping, (c) mixture experiments, and (d) final rewards. In blue is the baseline, in red experiments with Random IM, in orange with Regular IM and in green those with Improvement IM. All experiments were performed with 150 episodes and 25 runs, the plots show the current best rewards with 95\% confidence interval.}
    \label{fig:Cartpole}
\end{figure*}
\subsection{Cartpole Balancing}

Fig. 2 illustrates the results of the Cartpole experiments. The baseline learning curve shows very well the difficulties of BO with higher dimensional input spaces, in this case 15 dimensions. The mean learning rate breaks down after 20 episodes and cannot keep up with any of the IBO results beyond 35 episodes in the overall run time, with the exception of the regular-IM IBO Shaping experiment, Fig. 2b.

Comparing the learning rates of the IBO Experiments, shows that the Preference Results, Fig. 2a, have the least improvement of the mean but reduces the variance the most. In the shaping experiments, we noticed that IBO with pure shaping of parameters does not optimize without human input. This finding is especially noticeable in the experiments with the regular interactive metric, since from the 55th episode on there is no improvement away from the interaction episodes. Both the mean and the variance jump staircase-like every 25 episodes. This fact led us to develop PEI. In comparison, in the Mixture experiments,Fig. 2c, the big jumps happen through the user input, but BO manages to increase the reward between the interactions slightly and to reduce the variance. 

In Fig. 2d, the final rewards can be observed. IBO can outperform the baseline in all experiments. It is remarkable that the mixture experiments perform almost identically in both the mean and the variance, independent of the IM. This suggests that it is not important when and how regularly an interaction occurs, but that both the preference and the modification of the parameters are important. More results can be found in the supplementary pages.
\begin{figure*}
    \centering
    \includegraphics[width=\textwidth]{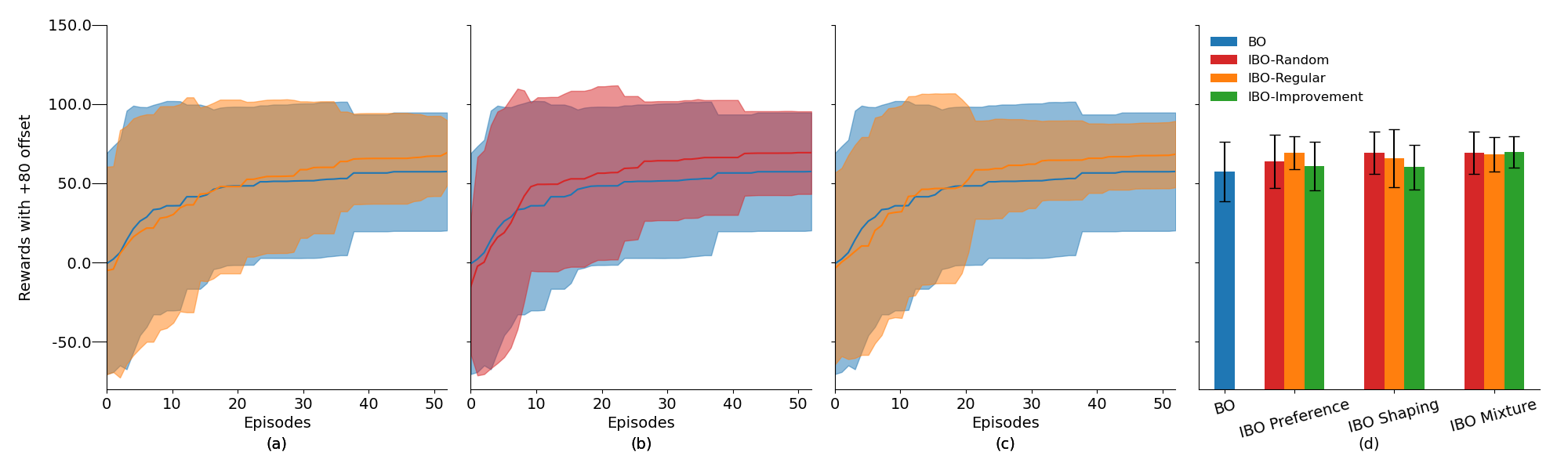}
    \caption{Shows Reacher results for (a) preference, (b) shaping, (c) mixture experiments, and (d) final rewards. In blue is the baseline, in red experiments with Random IM, in orange with Regular IM and in green those with Improvement IM. All experiments were performed with 50 episodes and 25 runs, the plots show the current best rewards with 95\% confidence interval.}
    \label{fig:Reacher}
\end{figure*}
\subsection{Reacher}
The Reacher experiments are intended to show two things, first, is it possible in principle to use the framework for more dimensional action spaces and what are its limitations? Second, how efficiently can a human interact with a reinforcement learning environment, what are the implications of a non-intuitive action space for humans?

As shown in Figure 3a-c, we can use our methods in the Reacher environment. It should be noted that in the course of the experiments the hard variant was chosen, which has a much smaller target. The reward function was changed so that each time step costs $-1$ and a reward of $10$ is granted when the target is reached. Furthermore, each action has a cost of $0.1|a|$, so  unnecessarily high energy consumption is penalized. 

As mentioned at the beginning, Reacher has an action space that is not intuitive for humans. It controls the joint accelerations, which is difficult for the user to imagine. This circumstance leads during the experiments to the fact that the direction in which Reacher moves is correct, but often the target is missed by just a little. Overall, the baseline is already hard to outperform, so iterations where the human just misses the target are a disadvantage for performance. However, as seen in all three results, Fig. 3a-c, all variants can improve the performance and reduce the variance of the results.

Fig. 3d shows again that the final results for IBO mixture performs almost the same regardless of IM. Furthermore, the figure shows how small the performance gap between baseline and IBO is. This can be explained by the rather 
simple experiment, whereby the baseline has no great difficulty in finding the optimal solution. Furthermore, as mentioned above, the user cannot develop a good intuition for the action space. Compared to Cartpole, Reacher also uses a sparse reward, which causes difficulties for BO, especially when no policy in the run has reached the goal yet.

\subsection{Robotic Task}
The last experiment demonstrates the usability of IBO on real world tasks. Here the Franka Emika Panda robot learns a two dimensional trajectory in Cartesian coordinate system, which is an optimal reaching policy. The limitation to two dimensions is due to the fact that we used the user interface from the Reacher experiment. We use a reward function that resembles that of the Reacher. The robot should reach its destination as quickly as possible. In addition, there is a terminate state if the robot enters an square shaped exclusion zone between start and end , which penalizes the agent with $- 100$. Since we don't have access to the real acceleration of the trajectory we don't optimize for energy efficiency. 
\begin{figure}
    \centering
    \includegraphics[scale=0.4]{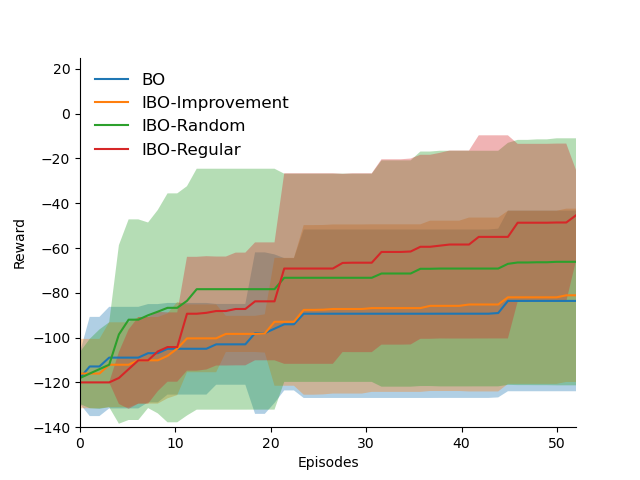}
    \caption{Shows Robotic experiment results. In blue is the baseline, in orange an Preference experiment with the Improvement IM was performed. The green denotes the Random IM experiment with Shaping, and in red the Regular IM was used for a Mixture experiment. All experiments were performed with 50 episodes and 10 runs, the plots show the current best rewards with 95\% confidence interval.}
    \label{fig:Franka}
\end{figure}

 We can outperform the baseline with IBO, although Preference is only marginally better. Due to the Cartesian action space, interaction with this environment is much easier. In order to ensure a comparable performance in the evaluations, we have preferred a maximum of two dimensions and modified two parameters during each interaction. 

In retrospect, we are convinced that a different representation model for the robotics experiment would be advantageous for the robot experiment, since due to the Gaussian basis it comes to unnecessary movements. A possible solution are Dynamic Movement Primitives \cite{a27} or Probabilistic Movement Primitives \cite{a28}.

\section{CONCLUSIONS}
Based on our experimental results, we have reasoned that our approach can indeed improve optimization with BO through human interaction. By continuously reducing the search space, Bayesian Optimization can continue to optimize in later episodes and thus makes possible the utilization of higher dimensional parametrization of the policy. Furthermore, we demonstrated with our framework that the comprehensible representation of higher dimensional parameter spaces and a simple interaction is possible. 

Due to the modular structure of our framework, the individual components can now be further developed and the application range and performance can be improved. Especially for action spaces with three or more dimensions a better way of display and interaction interface has to be developed, because of the confusing visualization on a display. Furthermore, we believe that it is a more accessible variant for robot applications not to display the actions but to use movement primitives to display the movements directly since humans have a higher understanding of them.

\addtolength{\textheight}{0cm}   



\section*{APPENDIX}

\section*{ACKNOWLEDGMENT}
This project has received funding from the Deutsche Forschungsgemeinschaft (DFG, German Research Foundation) - No \#430054590 (TRAIN).


\end{document}